\documentclass[twoside,11pt]{article}
\usepackage{xcolor}
\usepackage{blindtext}
\usepackage{tabularx} 
\usepackage{amsmath}
\usepackage{amssymb}
\usepackage{amsfonts}
\usepackage[utf8]{inputenc}
\usepackage[most]{tcolorbox}
\tcbuselibrary{breakable}
\usepackage{multirow}

\usepackage[abbrvbib, preprint]{jmlr2e}

\usepackage{jmlr2e}

\usepackage{lastpage}
\jmlrheading{xx}{2025}{1-\pageref{LastPage}}{2/25; Revised xx/25}{xx/25}{21-0000}{Javier Coronado-Blázquez}

\ShortHeadings{Book summaries from LLM internal knowledge}{Coronado-Blázquez}
\firstpageno{1}

\begin{document}

\title{Evaluating book summaries from internal knowledge in Large Language Models: a cross-model and semantic consistency approach}

\author{\name Javier Coronado-Blázquez \email j.coronado.blazquez@gmail.com \\
       \addr Telefónica Tech, AI \& Data Unit\\
       Madrid, 28050, Spain}

\editor{My editor}

\maketitle

\begin{abstract}
We study the ability of large language models (LLMs) to generate comprehensive and accurate book summaries solely from their internal knowledge, without recourse to the original text. Employing a diverse set of books and multiple LLM architectures, we examine whether these models can synthesize meaningful narratives that align with established human interpretations. Evaluation is performed with a LLM-as-a-judge paradigm: each AI-generated summary is compared against a high-quality, human-written summary via a cross-model assessment, where all participating LLMs evaluate not only their own outputs but also those produced by others. This methodology enables the identification of potential biases, such as the proclivity for models to favor their own summarization style over others. In addition, alignment between the human-crafted and LLM-generated summaries is quantified using ROUGE and BERTScore metrics, assessing the depth of grammatical and semantic correspondence. The results reveal nuanced variations in content representation and stylistic preferences among the models, highlighting both strengths and limitations inherent in relying on internal knowledge for summarization tasks. These findings contribute to a deeper understanding of LLM internal encodings of factual information and the dynamics of cross-model evaluation, with implications for the development of more robust natural language generative systems.\newline
\end{abstract}


\section{Introduction}
\label{sec:intro}

Large Language Models (LLMs) have become a standard tool for natural language processing (NLP) by generating high-quality text based on transformers (\cite{vaswani2023attentionneed}). LMs are deep neural networks that leverage this architecture to perform a wide range of natural language tasks. These models are trained on a vast corpus of text, learning to predict the next token in a sequence.

One of the most frequent use case for LLMs is summarization\footnote{Summarization is considered a task better suited for encoder-decoder architectures, as these models explicitly process the full input through an encoder before generating a compressed version via the decoder. However, in practice, decoder-only LLMs can perform summarization effectively without requiring an explicit encoding step. This is because they can condition the generation process on the input prompt, leveraging mechanisms like attention to retain and condense relevant information (\cite{fu2023decoderonlyencoderdecoderinterpretinglanguage,khandelwal2019sampleefficienttextsummarization}).}, i.e., condensing a text to its essential points and main arguments, producing a concise version that retains the original's core information. But, what happens when we ask an LLM to summarize a text it has no direct access to, but knows due to its training?

In this paper, we will prompt different LLMs to write a detailed summary of well-known books without the actual text or access to the internet. This way, the generated summary will be a product of its internal knowledge solely, enabling the evaluation of it and offering insights into the process of memorization by such models.

There has been several legal claims against most LLM developers for using copyrighted materials (including, but not limited to, literature works) as training data (\cite{li2024diggerdetectingcopyrightcontent,liu2024shieldevaluationdefensestrategies}). While the books tested in this work may have been part of training corpora, we will test whether these models are capable of generating fully detailed summaries, with timeline, characters, main events, etc. and compare them to reference, human-written summaries. Even if an actual book has not been seen by the model in the training stage, we use well-known books so that part of the training text revolves around it (including summaries themselves), in order to measure the accuracy and detail of zero-shot generated summaries.

The field of automatic text summarization has undergone several paradigm shifts, from traditional statistical methods to neural networks, and now to LLMs (see e.g. \cite{zhang2024systematicsurveytextsummarization} for a comprehensive review). Modern LLM-based summarization typically employs either extractive methods (selecting important phrases directly) or abstractive methods (generating summaries by rephrasing content).

Contrary to common assumptions, instruction tuning --not model size-- is the key factor in zero-shot summarization capability (\cite{10_1162_tacl_a_00632}). Their human evaluation revealed that even a 350M parameter instruction-tuned model can perform comparably to the 175B parameter GPT-3 model.

Also, summaries generated by LLMs are often preferred by human annotators over original reference summaries in standard datasets (\cite{liu2024learningsummarizelargelanguage}). This has led to an ``LLM-as-reference'' learning paradigm, where smaller summarization models are trained using LLM-generated summaries as targets.

Summaries are generally performed with access to the full target text (where limitations such as the lost-in-the-middle problem (\cite{liu2023lostmiddlelanguagemodels}) are found for very long texts). In this work, we are interested on book internal knowledge-based summaries, asking the model to summarize a well-known book without the actual text, just relying on its own knowledge.

LLMs have demonstrated remarkable capabilities to store and retrieve factual information. Yet, how this fact-based knowledge is actually encoded and stored within the parameters of these models remains an open question.

One of the most popular explanations is the ``Superposition Theory'' (\cite{elhage2022superposition}). Superposition describes the ability of neural networks to represent more features than they have dimensions by allowing features to be encoded in overlapping ways within the same parameter space.

This theory is based on the Johnson-Lindenstrauss (JL) Lemma (\cite{Johnson1984}), a mathematical result describing how high-dimensional data can be projected into a much lower-dimensional space while approximately preserving distances between points. This theoretical foundation provides insight into how LLMs can efficiently encode vast amounts of factual information within a finite number of parameters.

The JL lemma enables the so-called ``poly-semantic'' representations, where a single neuron can participate in representing multiple unrelated features simultaneously, as opposed to ``mono-semantic'' representations where each neuron corresponds to a single feature. This property allows neural networks to store exponentially more information than their raw parameter count would suggest.

Research at Anthropic using toy models has demonstrated that neural networks can not only store additional features in superposition by tolerating some interference but can also perform computations while in superposition (\cite{elhage2022superposition}). This suggests that LLMs may be ``noisily simulating'' larger, highly sparse networks within their more compact parameter space.

An alternative approach to explain this factual information memorization is binary encoding (\cite{tailcalled2024binary}). Here, features are embedded in a vector space by translating their indices into binary representations. By normalizing these binary vectors, it's possible to create embeddings where distinct features maintain relatively low similarity, enabling efficient storage of exponentially many features in a given vector space.

Regarding the types of memorization, there are usually two types distinguished: verbatim memorization and approximate memorization (\cite{kiyomaru-etal-2024-comprehensive-analysis}). Verbatim memorization refers to the literal reproduction of text from the training corpus, while approximate memorization allows for slight variations while preserving essential meaning.

The standard methodology for investigating memorization quantitatively involves providing an LLM with a prompt and analyzing whether its continuation replicates exact text from the training corpus. This approach has been instrumental in understanding how and when LLMs memorize versus generate novel content.

Several key factors influence LLMs memorization capabilities; most notably, parameter size, as larger models are expected to memorize more data than smaller ones. Also, longer prompts (context length) increase the probability of memorization. By simple duplication count, text that appears frequently in the training corpus is more likely to be memorized. Finally, text not included in the latter stages of training is less likely to be memorized, even if it appears frequently in the overall training corpus (\cite{menta2025analyzingmemorizationlargelanguage}).

Indeed, recent research has uncovered systematic patterns in how fact memorization scales with model size and training duration. \cite{lu-etal-2024-scaling} found that the LLM fact knowledge capacity has a linear relationship with model size and a negative exponential relationship with training epochs. These findings suggest that memorizing the entirety of a knowledge base like Wikidata would require training an LLM with approximately 1000 billion non-embedding parameters for 100 epochs, indicating that complete factual memorization may be impractical within current training paradigms.

This work also details how LLMs struggle with memorizing redundant facts in a unified way, showing inefficiency in handling correlated information. Thus, LLMs tend to prioritize memorizing frequent and difficult facts, with subsequent facts potentially overwriting prior memorization, which can hinder retention of low-frequency information.

\cite{chang2024how} provide insights into how LLMs acquire factual knowledge during the pre-training phase: acquisition occurs through a process of progressively increasing the probability of fact-based information presented in the training data at each step, though this increase can be diluted by subsequent forgetting. Counterintuitively, they found that training on more data doesn't necessarily improve a model's capability to acquire and maintain such knowledge. Models undergo both forgetting of memorization and generalization of factual data, with models trained on duplicated data exhibiting faster forgetting.

Additionally, training with larger batch sizes can enhance models' robustness to forgetting. These findings help explain observed behaviors of LLMs, such as poor performance on long-tail knowledge and the benefits of deduplicating training corpora.

In order to address the challenge of hallucination in LLMs, \cite{lin2024flamefactualityawarealignmentlarge} developed FLAME (FactuaLity-aware AlignMEnt), a procedure for enhancing the factual accuracy of LLMs during alignment. Their research identified key factors that lead to hallucination during both supervised fine-tuning (SFT) and reinforcement learning (RL) phases of alignment.

They found that training LLMs on unfamiliar knowledge encourages hallucination, making standard SFT less factual since it often includes novel information. Additionally, reward functions used in standard RL often inadequately capture factuality and favor longer, more detailed responses, inadvertently promoting hallucination. FLAME incorporates factuality awareness into both SFT and RL through direct preference optimization, guiding LLMs to produce more factual responses while maintaining their instruction-following capabilities.

The factuality of a LLM must be carefully evaluated, both on accuracy and consistency (see e.g, \cite{luo2023systematic,wang2023surveyfactualitylargelanguage}). Yet, it has been noted that LLMs tend to rate their own outputs more favorably than those generated by other models. This ``self-preference bias'' raises important questions about the reliability of using the same model as both generator and evaluator (\cite{zheng2023judgingllmasajudgemtbenchchatbot,xu2024prideprejudicellmamplifies}).

\cite{panickssery2024llm} conducted a study exploring whether LLMs can recognize their own outputs and how this capability affects evaluation bias, proving that models like GPT-4 and Llama 2 had ``non-trivial accuracy'' at distinguishing their own generated text from that of other models. Through controlled experiments, they established a linear correlation between self-recognition capability and self-preference bias, suggesting a causal relationship between these phenomena.

This research revealed that when acting as evaluators, LLMs consistently assign higher scores to outputs they themselves generated compared to equivalent-quality outputs from other models, a bias that contradicts human evaluator judgments. This finding has crucial implications for benchmarking methodologies that use the same model as both generator and evaluator.

\cite{wataoka2024selfpreferencebiasllmasajudge} introduced a novel quantitative metric to measure self-preference bias. Their research explored potential causes of this bias, particularly investigating whether LLMs favor outputs with lower perplexity (texts that are more predictable to them). Their findings indicate that LLMs assign significantly higher evaluations to outputs with lower perplexity than human evaluators do, regardless of whether the outputs were self-generated. This suggests the bias stems from familiarity with text patterns rather than just recognition of one's own outputs, adding nuance to our understanding of evaluation biases in LLMs.

In order to mitigate this phenomenon, we will use every generator model also as judge, performing blind cross-evaluation over the same generated summaries (where each judge model has no information on which model wrote the summary). This way, we can not only obtain averaged scores, but study the effect of self-preference bias in this experiment.

Additionally, we will use well-established metrics for NLP summarization tasks such as ROUGE (\cite{lin-2004-rouge}) and BERTScore (\cite{zhang2020bertscoreevaluatingtextgeneration}), which aim for n-gram and semantic similarity between reference. Nevertheless, these metrics may result insufficient for detailed and complex summaries, as small deviations in the text (leaving these scores almost identical) may result in factually incorrect information from the plot or character development.

The structure of this paper is as follows: in Section \ref{sec:methodology}, we detail the methodology of the study. Section \ref{sec:results} is devoted to the results and their analysis. We conclude in Section \ref{sec:conclusions}, highlighting key insights and outlining future work.

\section{Methodology}
\label{sec:methodology}

We start by elaborating a list of books from which LLMs will generate an internal knowledge-based summary. The selection of books, detailed in Appendix \ref{app:app_1}, attempts to have a heterogeneous sample regarding year, genre, extension, nationality, etc.

In order to establish a ground truth for comparison, we gather high-quality, publicly available human summaries from \href{https://www.sparknotes.com/}{Sparknotes}. The ratio between summary number of words and their corresponding book number of pages goes from $\sim0.75-1.0$ (Atlas Shrugged, Don Quixote, The Count of Monte Cristo, Gone With the Wind) up to $\sim6-10$ (The Catcher in the Rye, The Little Prince, Six Characters in Search of an Author, A Doll's House)\footnote{The number of pages may vary for different editions, so we take the value from current ones when available.}.

As noted in \cite{askari2025assessingllmszeroshotabstractive}, the consistency of LLMs when asked to execute summarization tasks is unstable. Therefore, to overcome this limitation, we perform five calls for each book and generator, in order to have some variance across them, focusing on the average scores.

A detailed prompt engineering is needed for an LLM to guarantee the task is performed as accurately as possible. The generator model is prompted the following instructions:

\begin{tcolorbox}[
  breakable,
  colframe=black,
  colback=white,
  title={}
]
\texttt{Provide a very detailed summary of the plot for the book "\{title\}" by \{author\}.
The summary must be of the original book, NOT any adaptation like a film or TV show.
Include all main events and the complete storyline, detailing every key development, situations, events with characters, and the conclusion.
Do NOT include any historical context, literary analysis, or philosophical discussion - only the plot.}
\end{tcolorbox}

Evaluation of the generated text by metrics such as ROUGE or BERTScore can be limited, as there are small variances in text which may lead to very different meanings or detail changes, while maintaining the scores almost identical. As a further evaluation, every generator model is used also as judge, for itself and for the rest, so we can evaluate the possible self-preference bias. Each generator model is only called once, to ensure that summaries evaluated by different judges are the same for each generator.

This LLM-as-a-judge evaluation is performed asking it to rate the summary in a 1--10 scale when comparing it to our reference text (ground truth). The prompt for this task is the following:

\begin{tcolorbox}[
  breakable,
  colframe=black,
  colback=white,
  title={}
]
\texttt{Compare the following generated summary with the original summary for the book.
Analyze whether all events, characters, and plot points are covered.
Provide a detailed justification of any missing or extra details.
At the end of your response, on a separate line, output ONLY a final numeric score (a single number between 1 and 10) for the generated summary based on the following criteria:
\newline
1 - The generated summary bears no meaningful relationship to the original.
\newline
2 - Only superficial or tangential connections exist.
\newline
3 - Identifies 1-2 basic elements but with major inaccuracies.
\newline
4 - Covers approximately 25\% of key content with significant errors or omissions.
\newline
5 - Addresses approximately 50\% of core elements but lacks nuance.
\newline
6 - Includes most major plot points but misses key details or characters.
\newline
7 - Covers all major events/characters but simplifies themes/subplots.
\newline
8 - Minor omissions in subplots or specific details.
\newline
9 - Near-perfect alignment with trivial discrepancies.
\newline
10 - Perfect parity in content, structure, and tone.
\newline
Ignore any external knowledge and focus solely on comparing the two summaries.
Provide your detailed justification and then the final score on a new line.}
\end{tcolorbox}

Finally, score extraction is performed with Gemini, and double-checked manually to ensure it is parsing the correct final score as a numeric value (as the format can change depending on the model and iteration). In this last step, the instructions passed to the model are:

\begin{tcolorbox}[
  breakable,
  colframe=black,
  colback=white,
  title={}
]
\texttt{Extract only the final numeric score (a number between 1 and 10) from the text below.
Output nothing but the number.
Text:}
\end{tcolorbox}

Six different LLMs are tested: Gemini 2.0 (Google), GPT-4o-mini (OpenAI), Llama 3.1-8b (Meta),  Mistral-7b (Mistral), Phi4-14b (Microsoft), and the 14b Qwen-distilled version of DeepSeek--R1 (DeepSeek)\footnote{In the case of DeepSeek--R1, as it outputs the full reasoning chain delimited by \texttt{<think>}, we first remove that part so that only the actual summary is passed to the judge model.}. These models provide variance of architectures and number of parameters to test different configurations on the same data set. All models are accessed locally via Ollama API and Python, with the exception of GPT and Gemini, which are used via their respective APIs.

After some preliminary tests, we set the temperature parameter of both summary generator and judge model to 0.4, as it is low enough to ensure an adherence to the factuality expected from a detailed book summary, yet high enough to allow the model to be creative in its generation and structure proposal\footnote{Very low temperatures of $\sim0.1-0.2$ are generally recommended for summarization tasks, in order to adhere to the factuality of the text. Yet, in this case the summary must be created from the internal knowledge of the model, requiring more ``imagination''.}.

\section{Results}
\label{sec:results}
\begin{figure}[!ht]
    \centering
    \includegraphics[width=0.8\linewidth]{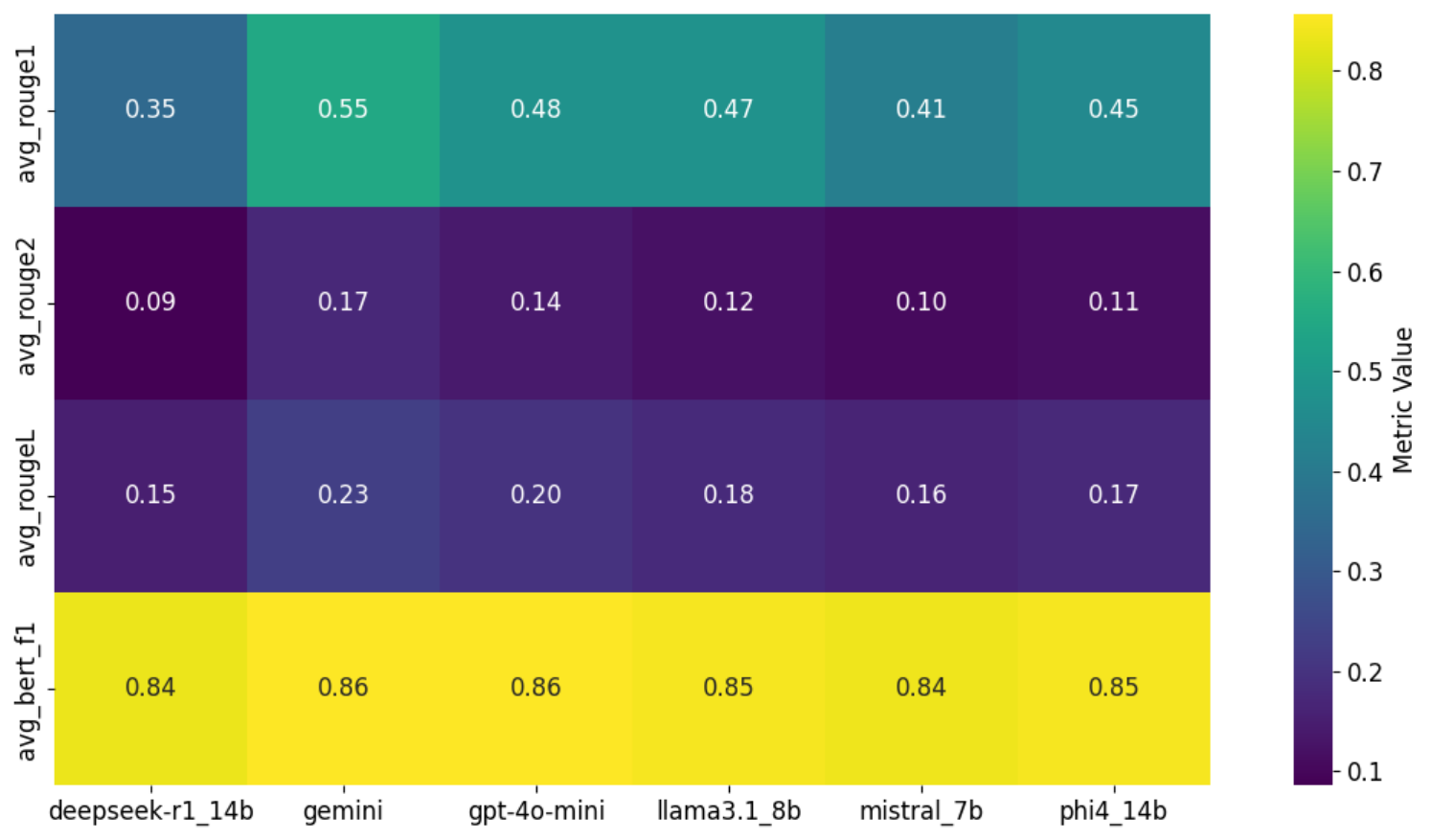}
    \caption{Heatmap of the ROUGE (ROUGE-1, ROUGE-2 and ROUGE-L) and BERTScore (F1) metrics, aggregated for the different LLMs.}
    \label{fig:rouge_bert_agg}
\end{figure}

The results are stored in individual files for their analysis. A first insight is to compute the metrics of ROUGE (including ROUGE-1, ROUGE-2 and ROUGE-L) and BERTScore (as F1) comparing generated vs. reference summaries, and aggregated for the different LLMs. This is shown in Figure \ref{fig:rouge_bert_agg} as a heatmap.

As depicted in the Figure, ROUGE-1 scores (measuring uni-grams) are around 0.5, while the score is much lower for ROUGE-2 (bi-grams) and ROUGE-L (for longest sub-sequences). When looking at the semantic similarity with BERTScore, we see values close to 0.9, with very little spread across different models. This indicates that every model understands the task very well and generate texts that have a lot in common (semantically) with the reference summary.

\begin{figure}[!ht]
    \centering
    \includegraphics[width=0.675\linewidth]{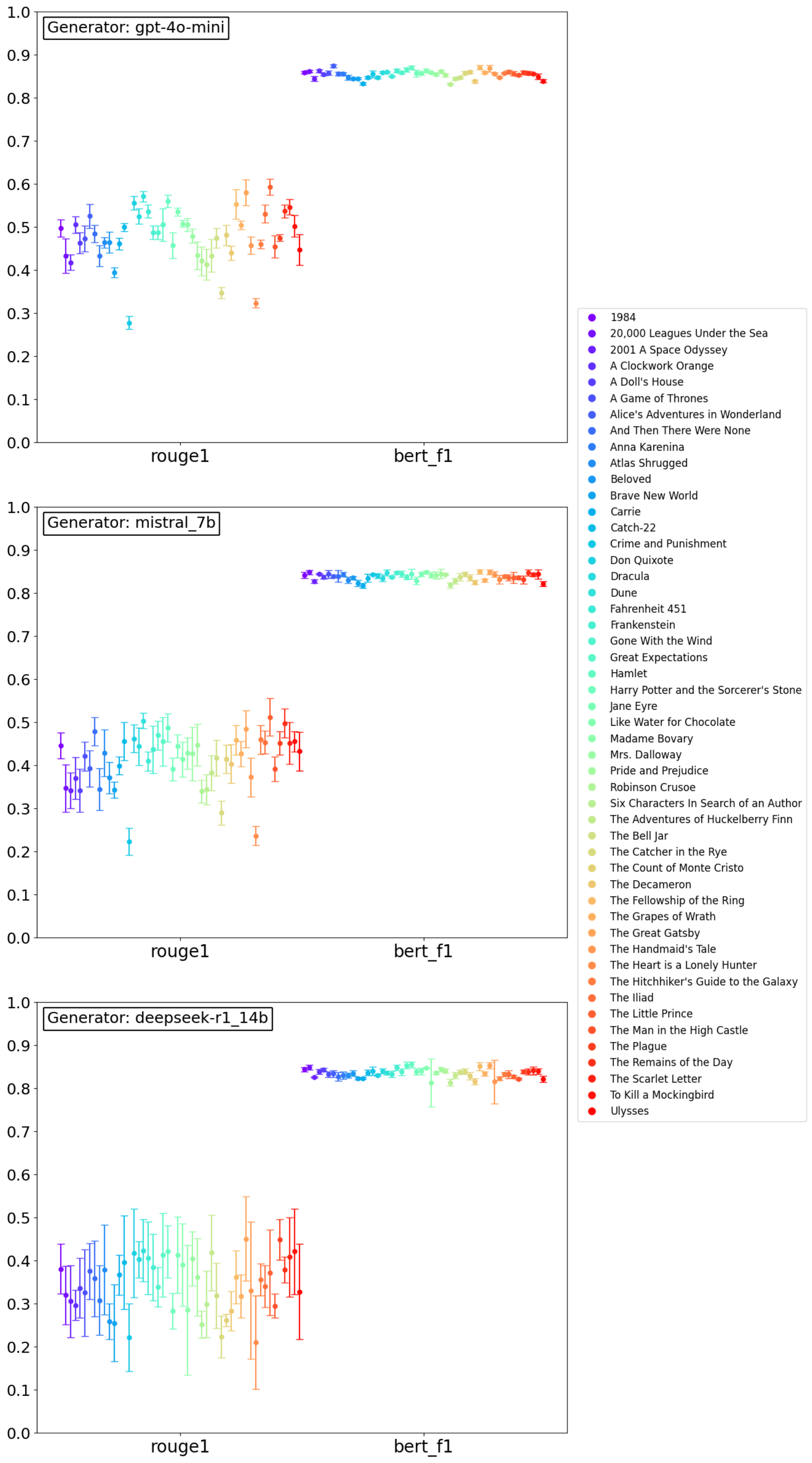}
    \caption{Comparison of ROUGE-1 and BERTScore metrics across the five calls per book and model. Error bars represent the 68\% confidence level. From top to bottom: results for GPT-4o-mini, Mistral-7b and DeepSeek-R1}
    \label{fig:rouge_bert_errorbars}
\end{figure}

The lack of consistency across different calls to a model can create large variance in these metrics for the same book. In Figure \ref{fig:rouge_bert_errorbars}, we plot the average score and 68\% confidence level error bars for two of these metrics (ROUGE-1 and BERTScore) when generated with three different models.

While the BERTScore values across the five repetitions of the same model and book are very consistent (indicating that the semantic similarity is preserved), the ROUGE-1 scores can be inconsistent, with error bars as large as $\sim50\%$ (revealing non-trivial differences in the uni-gram distribution), with diverse spread depending on the model: GPT-4o-mini presents more clustered values when compared to DeepSeek, which indicates more consistency in the uni-gram distribution (i.e., less grammatical diversity) across the five individual calls.

Yet, as mentioned before, semantic similarity alone may not be enough for evaluating such complex summaries, where specific details are crucial and relevant changes in how plot, descriptions or character developments are written may be invisible to those scores. To overcome this limitation, we use the LLM-as-a-judge paradigm, in which each model evaluates every model output individually (including its own).

This way, subtleties such as missing information about a event or a factually incorrect description of a character can be easily spotted and assessed by an LLM, which will take the rights and wrongs of the summary into consideration to evaluate the generated summary with a 1--10 score.

\begin{figure}[!ht]
    \centering
    \includegraphics[width=0.715\linewidth]{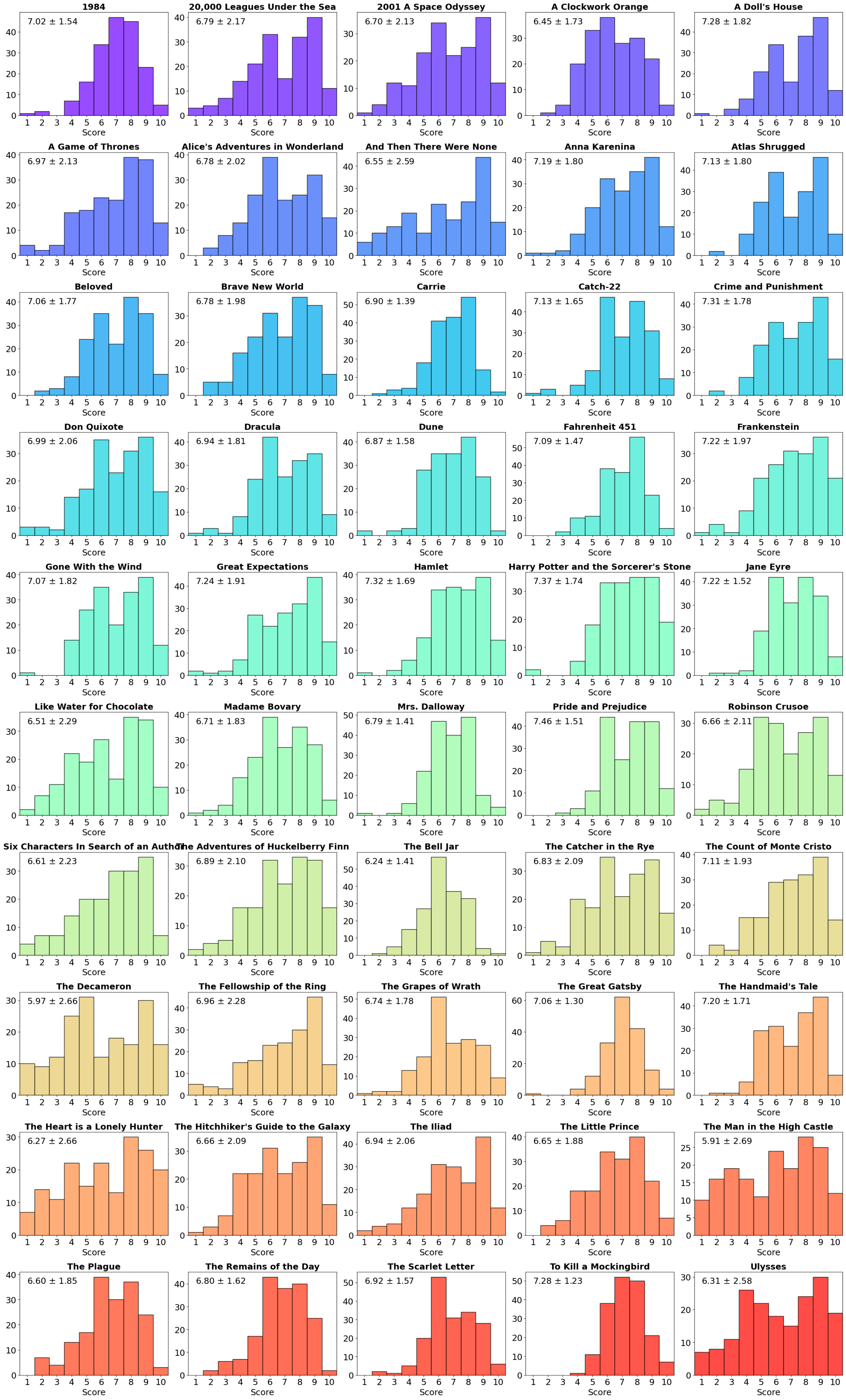}
    \caption{Distribution of LLM-based scores for all the books studied, in alphabetical order of title. Each histogram represents the distribution of all ratings in every combination of generator+judge. In the upper left corner of each subplot, the mean $\pm$ standard deviation of the distribution is shown.}
    \label{fig:scores_hist_all}
\end{figure}

We defer to Appendix \ref{app:app_2} some examples of low- and mid-score evaluations, where almost identical ROUGE or BERTScore values lead to very different judge scores. The overall distribution of scores for the different books is shown in Figure \ref{fig:scores_hist_all}

The mean values of the mosaic distributions indicate that, when aggregated by book, LLMs do not consistently write excellent summaries (at least, judged by LLMs themselves). Most of the average values are around 6-7, being the highest one 7.46 (for ``Pride and Prejudice''). It it interesting to observe the standard deviation, which is large for almost every case, around $20-30\%$, pointing towards low consistency both on generation and evaluation of the texts.

There is no pattern regarding a score bias depending on the book, this is, we do not observe significant differences by year, sex, genre or extension (science fiction or 19th century books are not better or worse summarized than adventure or 20th century ones). Likewise, there is no relation between the ratio of book length vs. summary extension and the quality of LLM-generated summary (long books shortly summarized and vice versa obtain similar scores). Thus, the internal knowledge of these well-known works of literature seems enough to tackle this task homogeneously, yet not excelling in average.

\begin{figure}[!ht]
    \centering
    \includegraphics[width=0.8\linewidth]{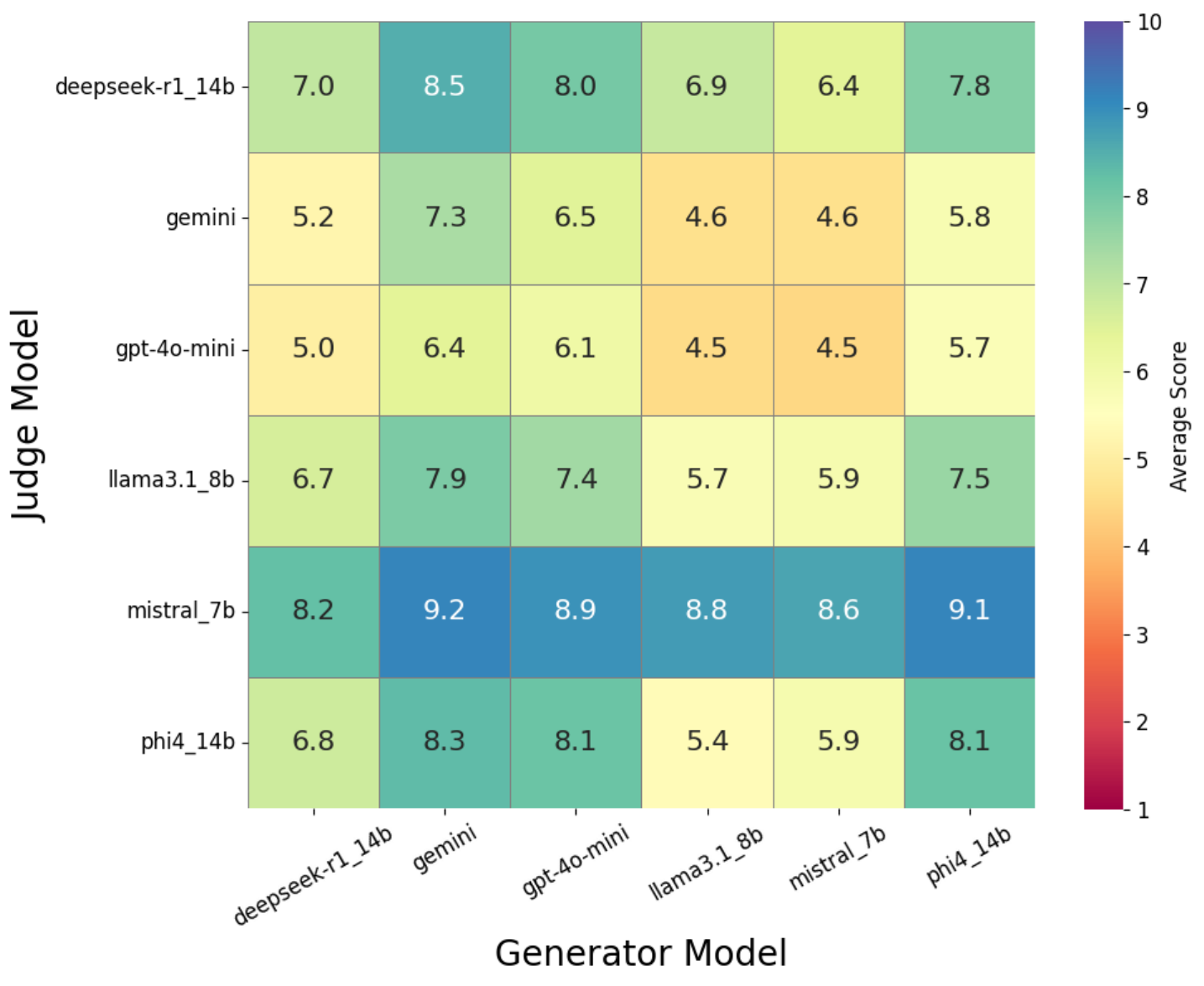}
    \caption{Heatmap of scores on all combinations of generator+judge, averaged for all books in the sample. Diagonal cells represent self-evaluated models.}
    \label{fig:heatmap_gen_judge}
\end{figure}

In order to understand the scores by model, both generator and judge, we show in Figure \ref{fig:heatmap_gen_judge} a heatmap with the average score for all the tested LLM combinations.

In this heatmap, the diagonal represent self-evaluated models, summaries generated and judged by the same model (without knowledge of it). As cited before, there is evidence that LLMs tend to assign higher scores when evaluating their own output, as if they recognized --and prefer-- their style. Yet, in this work we do not find evidence to support such bias, as there are several cases where a model is better scored by other LLM than by itself. The exception is Gemini 2.0, as its own output is rated 0.8 points higher (on average) than the second best. Being this the only case, we may conclude that these models are unable to distinguish and favor their own summaries.

The globally best generator models are Gemini 2.0 and GPT-4o-mini (which are the largest by number of parameters, although unknown), with average scores of $7.9\pm0.9$ and $7.5\pm1.1$ respectively. On the other hand, the worst ones are Llama 3.1-8b and Mistral-7b, both with $5.9\pm1.5$ average scores.

Interestingly, the two highest-scored generator models  are also the harshest judges, giving average scores of $5.4\pm1.1$ (Gemini 2.0) and $5.4\pm0.8$ (GPT-4o-mini). Mistral-7b, with an average score of $8.8\pm0.3$, is a non-reliable judge, as it systematically assigns every model with high ratings. Yet, we do not find any problem in its score justification; it simply does not give importance to aspects other models consider critical, or penalizes them with less points.

Additionally, it is worth noting the average ratings given by GPT-4o-mini and Gemini 2.0 when acting as judges are practically identical for every model (differing 0.2 points at most), except in their cross-evaluation, where they differ 0.4 and 0.9 points.

In the case of Llama 3.1-8b, in some cases it claims the generated summary is a verbatim copy from Wikipedia (happening for every model, including itself), assigning very low scores for considering it plagiarism. We double-checked and no similar structure or vocabulary, let alone literal chunks, were found in Wikipedia.

\begin{figure}[!ht]
    \centering
    \includegraphics[width=0.48\linewidth]{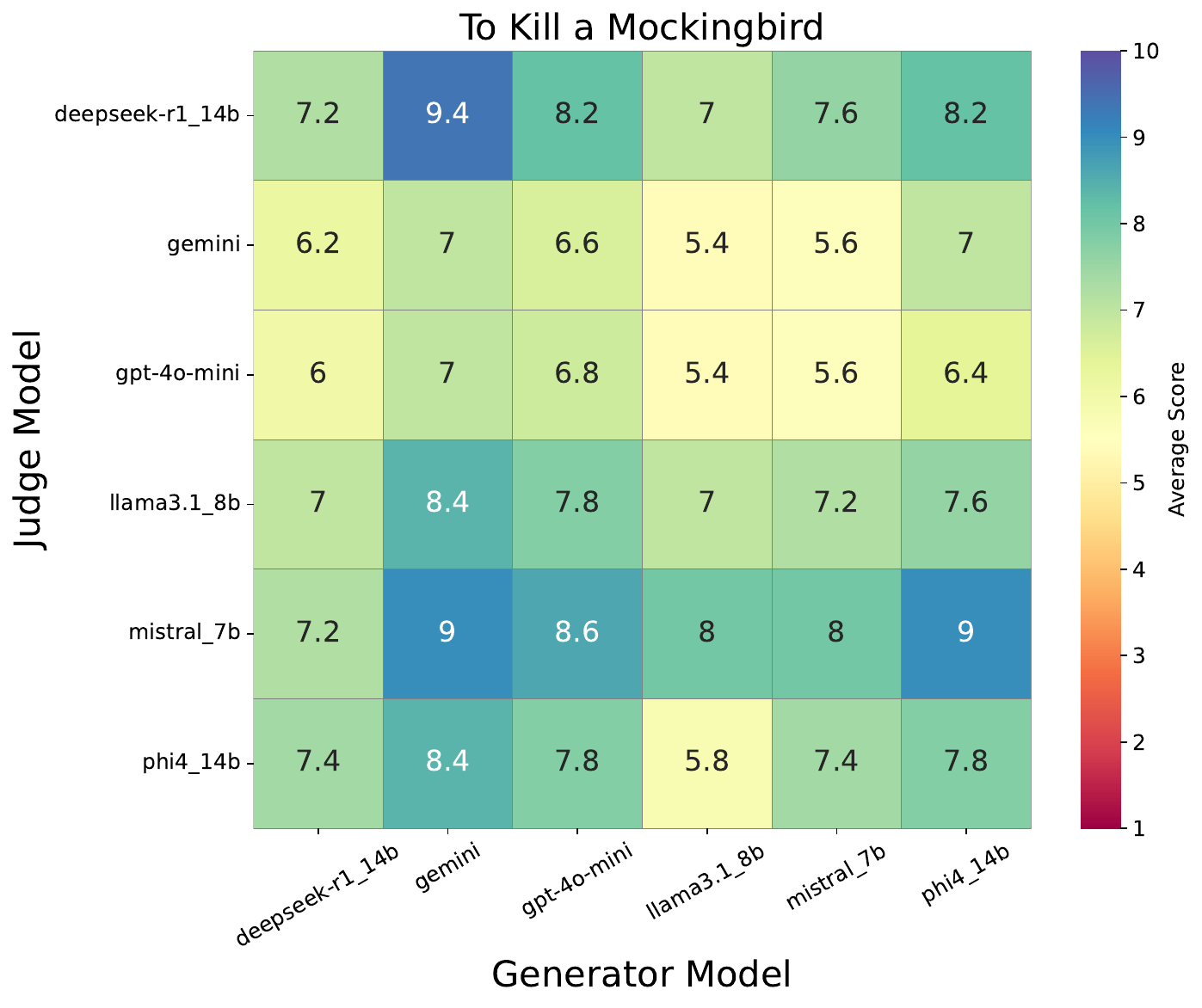}
    \vfill
    \includegraphics[width=0.48\linewidth]{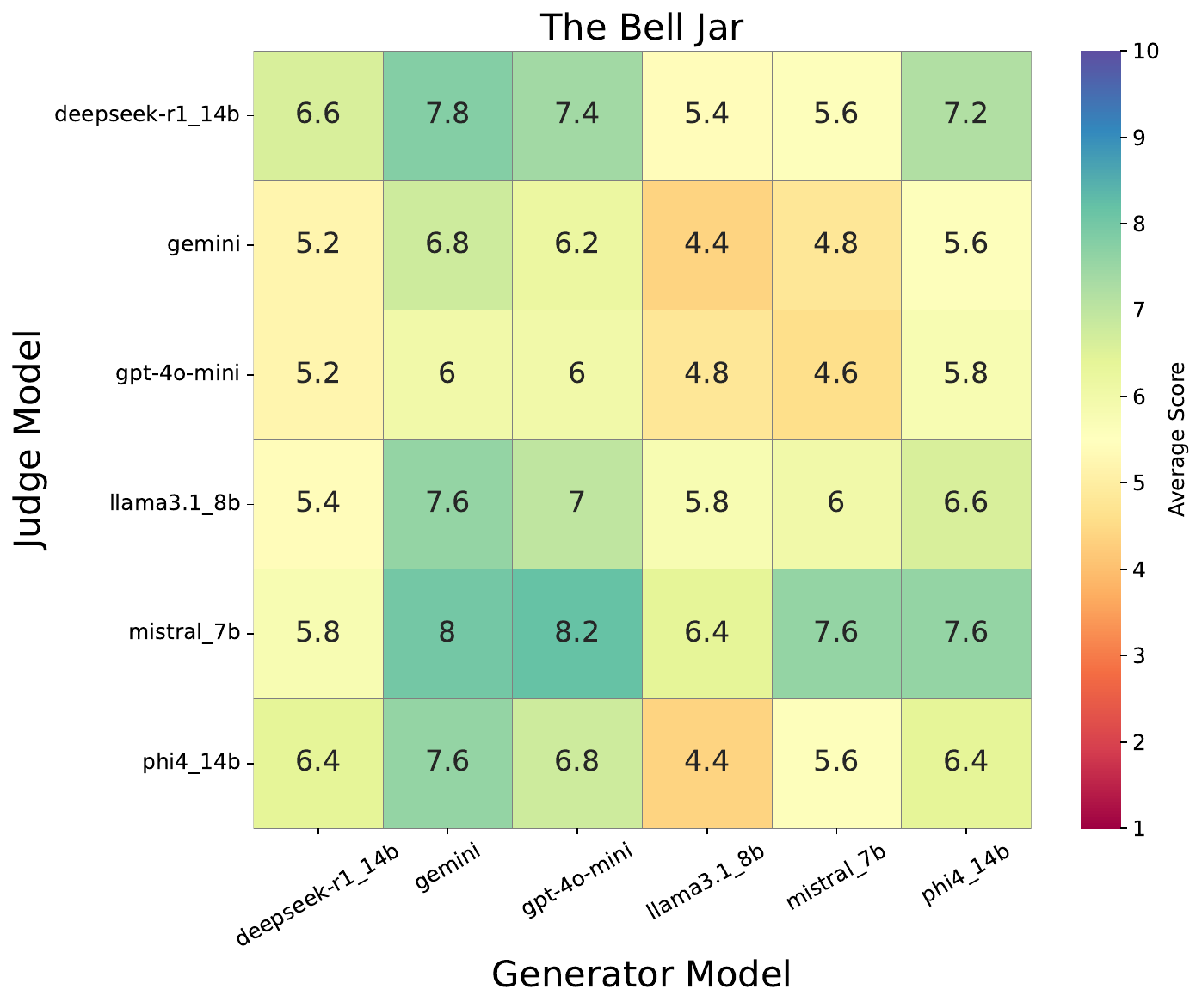}
    \vfill
    \includegraphics[width=0.48\linewidth]{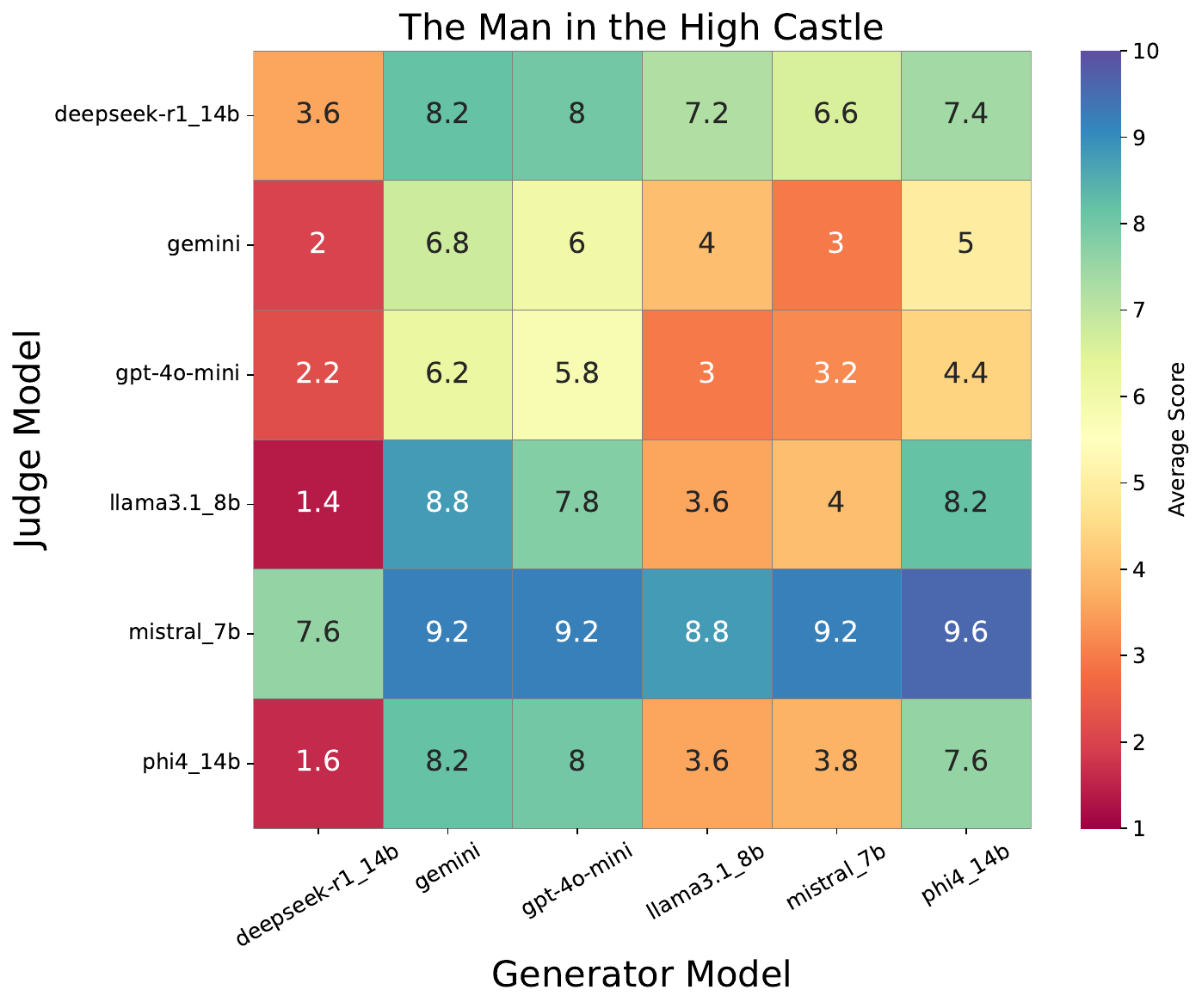}
    \caption{Heatmap of scores on all combinations of generator+judge, for three specific books. \textbf{Top panel:} To Kill a Mocking Bird. \textbf{Middle panel:} The Bell Jar. \textbf{Bottom panel:} The Man in the High Castle.}
    \label{fig:heatmap_three_books}
\end{figure}

As saw in Figure \ref{fig:scores_hist_all}, the distribution of scores can significantly differ depending on the book. In Figure \ref{fig:heatmap_three_books}, we show the heatmap of the generator+judge combinations for three illustrative examples.

These results point towards relevant differences of a summary quality depending on the book, the generator and evaluator model. In these three selected examples, we see a more or less homogeneous --low standard deviation-- and well-rated book (To Kill a Mockingbird, $7.3\pm1.2$ score), a still homogeneous but lower scored (The Bell Jar, $6.2\pm1.4$ rating, noting that even Mistral-7b does not systematically assign very high marks), and a very inhomogeneous one (The Man in the High Castle, $5.9\pm2.7$ score), where models with almost perfect scores coexist with almost null ratings.

This suggests that model knowledge of these books can be limited when devoting to such tasks, as it may be insufficient to provide enough detail on specific events of a complex work, especially for some models. In this sense, ``The Man in the High Castle'' seems to be worse memorized through the model internal encodings than ``To Kill a Mockingbird'', as its summaries are less thorough and detailed, miss relevant pieces of information or misplace events and characters.

Thus, the representation of topics concerning a well-scored book is expected to be redundant, this is, the model will have seen them many times in different contexts during its training. Likewise, when distances and semantic relationships between a book’s concepts are preserved, the model can extract and group the relevant elements coherently. In this case, each component of the prompted book is mapped to well-defined regions in the latent space, facilitating precise retrieval and summary generation.

Conversely, within the superposition theory of representation, a poor summary results when such phenomenon dilutes the signal, mixing plot elements inadequately, or when essential relationships between them are not preserved, reflecting a less robust and more diffuse internal representation.

\begin{figure}[!ht]
    \centering
    \includegraphics[width=0.65\linewidth]{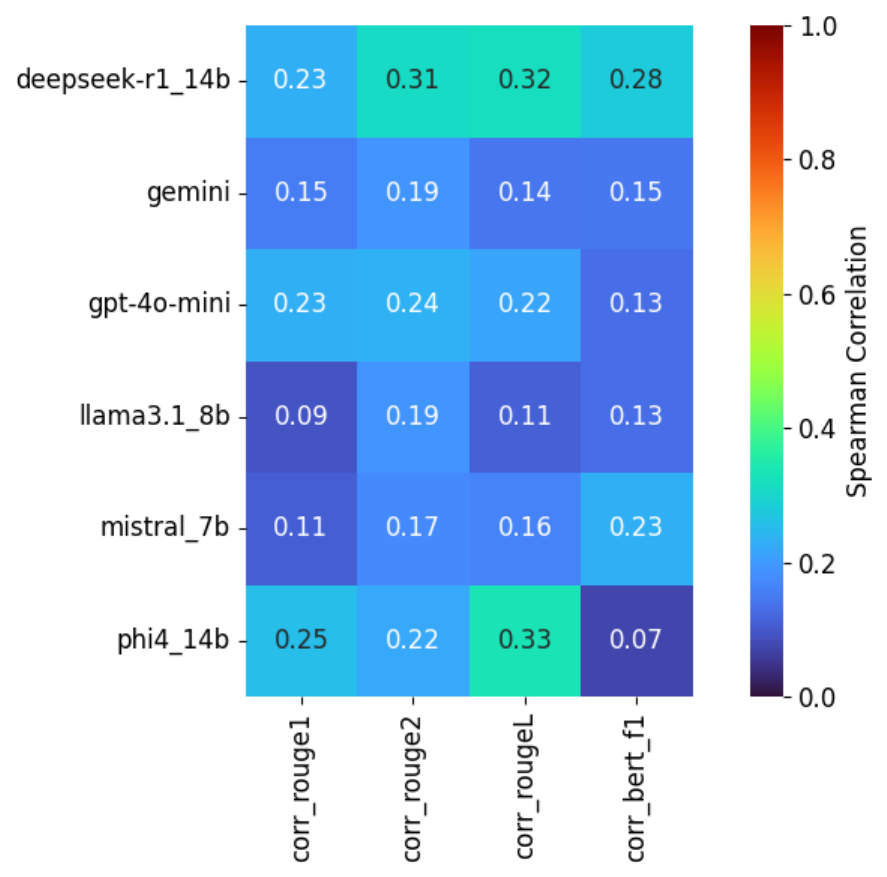}
    \caption{Spearman correlation between the LLM-as-a-judge average scores and the ROUGE and BERTScore metrics, for each of the generator models.}
    \label{fig:spearman_metrics}
\end{figure}

Finally, we test whether there is any relation between the LLM judge scores and the ROUGE/BERTScore metrics, by computing the Spearman correlation for each model individually. We show the results in Figure \ref{fig:spearman_metrics}.

We find no significant correlations between the different metrics and scores (although always positive), supporting our previous claim that these are not enough for evaluating very detailed summaries, where even slight omissions or details can very much alter their overall quality while not affecting ROUGE and/or BERTScore.

\section{Conclusions}
\label{sec:conclusions}

In this paper, we have studied the ability of Large Language Models (LLMs) to generate accurate and detailed summaries of well-known books from internal knowledge solely, i.e., without having the actual text available. We selected fifty books covering different eras, genres, nationalities and extensions, in order to have a diverse and heterogeneous sample to test.

Six LLMs were tested, with different number of parameters (ca. 7b, 14b and very large models accessible via API). As these models are known to present low consistency when prompted with such tasks, we repeated each individual call five times, so there is a statistical variance and we can focus on average values.

The evaluation of the generated summaries was performed by comparing them to benchmark, high-quality human summaries of the books in English language. These summaries were diverse in extension and, in particular, the ratio between the summary and book extension ranges more than one order of magnitude. Yet, no significant differences were found regarding this aspect.

This comparison with the benchmark summaries was done with two strategies: first, using standard summarization metrics such as ROUGE and BERTScore, in order to find overlaps in N-grams and semantic similarity. The results can vary depending on the model and book for the ROUGE metrics, yet present very little deviations for BERTScore (and all around $\sim0.85$). This indicates that the semantic concepts are very similar between reference and generated summaries, which proves LLMs perfectly understand this task and have sufficient internal knowledge to tackle it.

Yet, semantic similarity can be misleading, as deviations in specific details in plot key points, character developments or events timeline can significantly alter the meaning and factuality of the summary, severely impacting its overall quality. To overcome this limitation, we employed a second evaluation strategy, using the LLM-as-a-judge paradigm.

With this evaluation method, every model was used as generator but also as judge, for itself and for the rest. This way, we established a blind cross-evaluation between LLMs, to search for possible biases and limitations, such as the well-known ``self-preference bias'', where LLMs tend to favor their own outcomes.

The results reject this self-preference bias except for Gemini 2.0, as most models assign higher average scores to other models than to themselves. But the main insight in this LLM-as-a-judge evaluation is that these averaged ratings are much lower --and less consistent-- than the semantic similarities, supporting our previous claim regarding the limitation of this metric as evaluator. The models were asked to justify in detail the assigned score, where most models make minor to relevant errors in factual descriptions of the book elements, such as characters, plot events, timeline...

To conclude, although LLMs have indeed the internal knowledge of these renowned works of literature, they prove to be unable to systematically generate excellent, human-quality summaries from zero-shot prompting, being capable of just writing (on average) mediocre ones, with hallucinations, confusion between situations, omission of details and factual errors.

Future work may extend linguistic coverage, as these models present different biases depending on the prompted language (see, e.g \cite{coronadoblázquez2025deterministicprobabilisticpsychologyllms} for the effect of language on random number generation bias with LLMs). Additionally, in this work we have tested full summaries, yet it could be interesting to study the ability of these models to summarize only parts of books, this is, a ``spoiler-free'' summary up to a certain part (e.g, 50\%, 70\%), while contrasting it with cut versions.

\section*{Acknowledgements}
The author thanks Pedro Reviriego and Javier Conde for fruitful and enriching discussions leading to this work.

\appendix
\section{Book selection}
\label{app:app_1}
In Table \ref{tab:book_list} we show the full, fifty-book sample analyzed in this work.

\begin{table}[!ht]
    \centering
    \footnotesize
    \begin{tabular}{c|c}
        \bf Author & \bf Title \\
        \hline
        \hline
        Douglas Adams & The Hitchhiker's Guide to the Galaxy \\
        Margaret Atwood & The Handmaid's Tale \\
        Jane Austen & Pride and Prejudice \\
        Giovanni Boccaccio & The Decameron \\
        Ray Bradbury & Fahrenheit 451 \\
        Charlotte Brönte & Jane Eyre \\
        Anthony Burgess & A Clockwork Orange \\
        Albert Camus & The Plague \\
        Lewis Carroll & Alice's Adventures in Wonderland \\
        Agatha Christie & And Then There Were None \\
        Arthur C. Clarke & 2001: A Space Odyssey \\
        Miguel de Cervantes & Don Quixote \\
        Charles Dickens & Great Expectations \\
        Daniel Defoe & Robinson Crusoe \\
        Philip K. Dick & The Man in the High Castle \\
        Fyodor Dostoyevsky & Crime and Punishment \\
        Alexandre Dumas & The Count of Monte Cristo \\
        Laura Esquivel & Like Water for Chocolate \\
        F. Scott Fitzgerald & The Great Gatsby \\
        Gustave Flaubert & Madame Bovary \\
        Nathaniel Hawthorne & The Scarlet Letter \\
        Joseph Heller & Catch-22 \\
        Frank Herbert & Dune \\
        Aldous Huxley & Brave New World \\
        Homer & The Iliad \\
        Henrik Ibsen & A Doll's House \\
        Kazuo Ishiguro & The Remains of the Day \\
        James Joyce & Ulysses \\
        Stephen King & Carrie \\
        Harper Lee & To Kill a Mockingbird \\
        George R. R. Martin & A Game of Thrones \\
        Carson McCullers & The Heart is a Lonely Hunter \\
        Margaret Mitchell & Gone With the Wind \\
        Toni Morrison & Beloved \\
        George Orwell & 1984 \\
        Sylvia Plath & The Bell Jar \\
        Luigi Pirandello & Six Characters In Search of an Author \\
        Ayn Rand & Atlas Shrugged \\
        J. K. Rowling & Harry Potter and the Sorcerer's Stone \\
        J.D. Salinger & The Catcher in the Rye \\
        William Shakespeare & Hamlet \\
        Mary Shelley & Frankenstein \\
        Bram Stoker & Dracula \\
        John Steinbeck & The Grapes of Wrath \\
        Antoine de Saint-Exupèry & The Little Prince \\
        J. R. R. Tolkien & The Fellowship of the Ring \\
        Lev Tolstoi & Anna Karenina \\
        Mark Twain & The Adventures of Huckleberry Finn \\
        Jules Verne & 20,000 Leagues Under the Sea \\
        Virginia Woolf & Mrs. Dalloway \\
    \end{tabular}
    \caption{Selection of books used in this work, in alphabetical order of author.}
    \label{tab:book_list}
\end{table}

\section{Examples of LLM-as-a-judge score justifications}
\label{app:app_2}

In this appendix, we show some examples of LLM evaluations. Although explicitly prompted to generate a summary from the original book, and not from any adaptation, there are some cases where a model bases the text upon a movie or TV show. In the example below, DeepSeek--R1 is clearly using the Amazon Original series of ``The Man in the High Castle'' instead of the Philip K. Dick book. We show the Llama 3.1-8b evaluation: 

\begin{tcolorbox}[
  breakable,
  colframe=black,
  colback=white,
  title={DeepSeek--R1 evaluated by Llama 3.1-8b, ``The Man in the High Castle''}
]
\texttt{
The generated summary bears little resemblance to the original. It appears to be a conflation of several science fiction elements, including time travel and Nazi plots, with no connection to the actual book ""The Man in the High Castle"" by Philip K. Dick.
\\
\newline
* The main characters are entirely different from those in the original summary:
\newline
	+ Frank Childan is replaced by ""Franklod Farnsworth,"" an amateur scientist who invents a time machine.
    \newline
	+ Robert Baynes is replaced by ""Joe Levenstein.""
    \newline
	+ Nobusuke Tagomi is renamed to ""Mr. Tagomi"" and his character traits are altered significantly.
    \newline
    \newline
* The plot summary is also incorrect:
\newline
	+ There is no mention of the film ""The Grasshopper Lies Heavy"" being a sought-after item or its significance in challenging the established order.
    \newline
	+ Time travel, which is not present in the original book, plays a central role in the generated summary.
    \newline
	+ The ending, which mentions history's mutability and multiple realities, is also unrelated to the actual conclusion of the book.
\\
\newline
The generated summary seems to be an amalgamation of elements from various science fiction stories, including ""The Man in the High Castle,"" but it does not accurately represent the original book.
\\
**Final Score:** 1/10}
\newline
\end{tcolorbox}

Gemini 2.0 is especially good in structuring its justification when acting as judge. An example is shown below, evaluating the output of GPT-4o-mini for ``Catch-22'' in a summary scored with 7/10:

\begin{tcolorbox}[
  breakable,
  colframe=black,
  colback=white,
  title={GPT-4o-mini evaluated by Gemini 2.0, ``Catch-22''}
]
\texttt{The generated summary is quite comprehensive, covering many of the key elements of the original summary and the book itself. However, there are some omissions and slight inaccuracies.
\\
\\
**Strengths:**
\\
*   **Core Plot:** The generated summary accurately captures the central plotline of Yossarian's desire to escape the war, his feigning illness, and his frustration with the bureaucratic absurdities.
\\
*   **Key Characters:** It identifies and briefly describes several important characters, including Colonel Cathcart, Milo Minderbinder, Doc Daneeka, Nately, and Orr.
\\
*   **Catch-22 Definition:** It provides a decent explanation of the Catch-22 concept and its paradoxical nature.
\\
*   **Ending:** It accurately describes Yossarian's decision to desert and flee to Sweden.
\\
*   **Themes:** It touches upon the themes of absurdity, dehumanization, and the individual's struggle against bureaucracy.
\\
\\
**Weaknesses/Omissions:**
\\
*   **Snowden:** The summary completely omits any mention of Snowden, whose death is a pivotal event influencing Yossarian's perspective and actions. This is a significant omission.
\\
*   **Mission Focus:** The original summary highlights the emphasis on aerial photographs over actual target destruction, which is a key element of the book's satire. The generated summary mentions bombing missions but doesn't emphasize this specific absurdity.
\\
*   **Nately's Girlfriend's Revenge:** The original summary mentions Nately's girlfriend's attempts to stab Yossarian after Nately's death. This is included in the generated summary, but it is not emphasized enough.
\\
*   **Milo's Syndicate Details:** While Milo's black-market empire is mentioned, the detail about ""everyone having a share"" and the falsity of that promise is not included.
\\
*   **Cathcart and Korn's Offer:** The generated summary doesn't mention the specific offer made by Cathcart and Korn (court-martial vs. honorable discharge with conditions).
\\
*   **Yossarian's Wanderings in Rome:** The summary mentions Yossarian being in Rome, but it doesn't detail his encounters with the horrors he witnesses there.
\\
*   **Tone:** While the generated summary captures some of the absurdity, it lacks the darkly humorous and satirical tone of the original summary and the book.
\\
\\
Overall, the generated summary covers most of the major plot points and characters, but it misses some key details and nuances, particularly the significance of Snowden and the specific absurdities that drive Yossarian's actions. Score: 7}
\end{tcolorbox}

As a final example, we show a summary of ``Brave New World'' by Llama 3.1-8b scored by Gemini 2.0. In this case, there are many inaccuracies and fabricated events, due to either confusion or hallucination of the model:

\begin{tcolorbox}[
  breakable,
  colframe=black,
  colback=white,
  title={Llama 3.1-8b evaluated by Gemini 2.0, ``Brave New World''}
]
\texttt{The generated summary has several inaccuracies and omissions when compared to the original summary.
\\
\\
**Inaccuracies:**
\\
*   **Bernard's Caste:** The generated summary states Bernard is an Alpha-Plus, the highest caste. The original summary implies he is an Alpha, but specifically notes that he is small and weak for his caste, which is a source of his discontent.
\\
*   **Lenina's Relationships:** The generated summary incorrectly states Lenina becomes infatuated with Henry Foster again and has an affair with Helmholtz Watson. The original summary shows her attraction to John and her confusion over his rejection.
\\
*   **Helmholtz's Rebellion:** The generated summary mentions Helmholtz's rebellion being limited to writing anti-conformist literature. The original summary states he is unhappy because he is too intelligent for his job writing hypnopaedic phrases, which is more specific.
\\
*   **John's Destination:** The generated summary incorrectly states John is brought back to London to study at the World State University. The original summary does not mention any such study.
\\*   **John's Love Life:** The generated summary incorrectly states Lenina is married to Henry Foster. The original summary shows John's love for Lenina and her confusion over his rejection.
\\*   **Bernard's Conformity:** The generated summary states Bernard eventually returns to his conformist ways. The original summary shows Bernard being exiled.
\\*   **Linda's Return:** The generated summary incorrectly states Linda was sent back to London by her family to give birth to Bernard. The original summary states she was lost during a storm on the reservation.
\\*   **John's Leadership:** The generated summary incorrectly states John is now the leader of the tribe. The original summary does not mention this.
\\*   **Ending:** The generated summary's ending is completely fabricated and bears no resemblance to the original summary's ending.
\\
\\
**Omissions:**
\\
*   **Hatching and Conditioning Processes:** The generated summary mentions Bokanov's Process but omits details about the Podsnap Process and the conditioning of embryos into different castes.
\\
*   **Hypnopaedia:** The generated summary does not mention hypnopaedia (sleep-teaching) and its role in instilling World State values.
\\
*   **Director's Story:** The generated summary omits the Director's story about losing a woman on the Reservation.
\\
*   **Shakespeare:** The generated summary omits the significance of Shakespeare in John's upbringing.
\\
*   **Soma:** The generated summary mentions soma but doesn't fully convey its role in controlling emotions and maintaining social stability.
\\
*   **Mustapha Mond:** The generated summary does not mention Mustapha Mond and his debates with John about the value of the World State.
\\
*   **John's Self-Flagellation:** The generated summary omits John's retreat to the lighthouse and his self-flagellation.
\\
*   **Orgy and Suicide:** The generated summary omits the orgy and John's subsequent suicide.
\\
\\
The generated summary captures some basic elements of the plot, such as the controlled society, Bernard's dissatisfaction, and John's rebellion. However, it contains significant inaccuracies and omits many key details, characters, and plot points. Score: 5/10}
\end{tcolorbox}

\vskip 0.2in
\bibliography{references}

\end{document}